\documentclass[11pt,letterpaper]{article}

\usepackage[utf8]{inputenc}
\usepackage[T1]{fontenc}
\usepackage[english]{babel}

\usepackage{lmodern} 
\usepackage{microtype} 
\usepackage{setspace} 
\usepackage[
    letterpaper,
    top=1in,
    bottom=1in,
    left=1in,
    right=1in,
    headheight=14pt
]{geometry}

\usepackage{amsmath,amssymb,amsfonts,amsthm}
\usepackage{mathtools} 

\usepackage{graphicx}
\usepackage{booktabs} 
\usepackage{array} 
\usepackage{multirow} 
\usepackage{subcaption} 
\usepackage[font=small,labelfont=bf]{caption} 
\usepackage{float} 

\usepackage{enumitem} 
\setlist[itemize]{topsep=0pt,itemsep=2pt,parsep=2pt}
\setlist[enumerate]{topsep=0pt,itemsep=2pt,parsep=2pt}

\usepackage{pifont} 
\usepackage{wasysym} 

\usepackage{fancyhdr}
\usepackage[
    backend=biber,
    style=numeric-comp,
    sorting=none,
    maxbibnames=10,
    minbibnames=10
]{biblatex}
\usepackage[dvipsnames]{xcolor}
\definecolor{linkblue}{RGB}{0,0,128}
\definecolor{citegreen}{RGB}{0,128,0}
\definecolor{urlred}{RGB}{128,0,0}

\usepackage{url}
\usepackage[
    colorlinks=true,
    linkcolor=linkblue,
    citecolor=citegreen,
    urlcolor=urlred,
    bookmarks=true,
    pdfborder={0 0 0},
    pdfauthor={Amir Hosseinian et al.},
    pdftitle={January Food Benchmark: A Public Benchmark Dataset for Multimodal Food Analysis},
    pdfsubject={Computer Vision, Food Recognition, Benchmark Dataset},
    pdfkeywords={food recognition, vision-language models, nutritional analysis, benchmark dataset}
]{hyperref}

\usepackage{authblk}

\newcommand{\eg}{\textit{e.g.}}

\usepackage{titlesec}
\titleformat{\section}
  {\normalfont\Large\bfseries}{\thesection}{1em}{}
\titleformat{\subsection}
  {\normalfont\large\bfseries}{\thesubsection}{1em}{}
\titleformat{\subsubsection}
  {\normalfont\normalsize\bfseries}{\thesubsubsection}{1em}{}
  
\titlespacing*{\section}{0pt}{3.5ex plus 1ex minus .2ex}{2.3ex plus .2ex}
\titlespacing*{\subsection}{0pt}{3.25ex plus 1ex minus .2ex}{1.5ex plus .2ex}

\renewenvironment{abstract}{%
    \small
    \begin{center}
    \bfseries \abstractname\vspace{-.5em}\vspace{0pt}
    \end{center}
    \list{}{%
        \setlength{\leftmargin}{0.5in}
        \setlength{\rightmargin}{\leftmargin}
    }
    \item\relax
}{\endlist}

\theoremstyle{plain}

\theoremstyle{definition}

\theoremstyle{remark}

\begin{document}

\title{\Large\textbf{January Food Benchmark (JFB):\\A Public Benchmark Dataset and Evaluation Suite\\for Multimodal Food Analysis}}

\author{Amir Hosseinian\thanks{Corresponding author (\href{mailto:amirhoss@january.ai}{ amirhoss@january.ai})}}
\author{Ashkan Dehghani Zahedani}
\author{Umer Mansoor}
\author{Noosheen Hashemi}
\author{Mark Woodward}

\affil{January AI}

\date{August 2025}

\thispagestyle{plain}

\maketitle

\begin{abstract}
Progress in AI for automated nutritional analysis is critically hampered by the lack of standardized evaluation methodologies and high-quality, real-world benchmark datasets. To address this, we introduce three primary contributions. First, we present the January Food Benchmark (JFB), a publicly available collection of 1,000 food images with human-validated annotations. The validation process followed a rigorous protocol to ensure high-quality, reliable ground-truth labels. Second, we detail a comprehensive benchmarking framework, including robust metrics and a novel, application-oriented overall score designed to assess model performance holistically. Third, we provide baseline results from both general-purpose Vision-Language Models (VLMs) and our own specialized model, \texttt{january/food-vision-v1}. Using this framework, we find that \texttt{january/food-vision-v1} establishes a strong baseline with an Overall Score of 86.2, a \textbf{12.1-point} improvement over the strongest general-purpose, oracle-prompt configuration (\texttt{GPT-4o (Best)}, 74.1). This work provides the research community with a valuable new evaluation dataset and a rigorous framework to guide and benchmark future developments in automated nutritional analysis.
\end{abstract}

\vspace{2ex}

\section{Introduction}

The rising prevalence of diet-related chronic diseases has underscored the urgent need for accessible and accurate dietary tracking tools. Automating food logging from images captured on mobile devices presents a user-friendly solution, yet the underlying AI task is fraught with difficulty~\cite{Schembre2021Appetite, Aguilar2023IEEESPM}. Evaluating the accuracy of an AI's output is a critical but challenging step. This difficulty stems from two core problems: the lack of standardized metrics and, more importantly, the absence of high-quality, publicly available datasets that can serve as a reliable ground truth for the nuanced domain of food analysis.

Existing general-purpose Vision-Language Models (VLMs) have demonstrated impressive capabilities across a wide range of tasks~\cite{Radford2021ICML, Liu2023NeurIPS}, though their performance on specialized, fine-grained domains remains an active area of investigation~\cite{Li2024arXiv}. The specific domain of food image analysis---which requires identifying a dish, recognizing its constituent ingredients, and estimating nutritional content---presents unique challenges. These include high visual similarity between different dishes, severe ingredient occlusion, and the need to infer non-visual attributes like portion size and cooking methods~\cite{Wei2022TPAMI, He2024arXivFoodVLM}. Without a specialized, high-fidelity dataset, it is difficult to quantify how well these general models perform in this domain or to measure the value of domain-specific fine-tuning.

To address these challenges, we present three primary contributions:

\begin{itemize}[leftmargin=*,topsep=0.5ex]
    \item \textbf{A High-Quality Benchmark Dataset (JFB):} We introduce the January Food Benchmark (JFB), a collection of 1,000 real-world food images with human validated annotations for meal names, ingredients, and macronutrients. The dataset is released under a CC-BY-4.0 license to encourage reproducibility and downstream research.

    \item \textbf{An Automated Benchmarking Framework:} We propose a comprehensive evaluation methodology with robust metrics for meal identification, ingredient recognition, nutritional estimation, latency, and cost. We also introduce a novel \textbf{Overall Score}, a weighted metric tailored to reflect application-oriented needs for a balanced, high-performing system.

    \item \textbf{Baseline Performance Scores:} We benchmark top general-purpose models, \textbf{evaluated under both average and best-case zero-shot conditions}, against our specialized model, \texttt{january/food-vision-v1}, to establish a quantitative performance ranking and quantify the gap between generalist and specialist approaches.
\end{itemize}

Our analysis quantifies the significant performance gains from a specialized approach, establishing a strong baseline for future research and development in this challenging domain.

\subsection{Comparison to Existing Resources}

While numerous food datasets exist, JFB is specifically designed to fill a gap in high-fidelity, meal-level analysis from unconstrained, real-world user photos. Table~\ref{tab:dataset_comparison} contrasts JFB with other prominent resources, highlighting its unique focus on validated, multi-faceted annotations for complex meals, coupled with an open evaluation suite.

\begin{table*}[ht!]
\centering
\caption{Comparison of JFB with existing food datasets and benchmarks.}
\label{tab:dataset_comparison}
\resizebox{\textwidth}{!}{%
\begin{tabular}{@{}lccccc@{}}
\toprule
\textbf{Dataset} & \textbf{Meal Name} & \textbf{Ingredient List} & \textbf{Macronutrients} & \textbf{Real-World Photos} & \textbf{Human Validation} \\
\midrule
Food-101~\cite{Bossard2014ECCV} & \checkmark & \ding{55} & \ding{55} & \ding{55} & \ding{55} \\
Recipe1M+~\cite{Marin2021TPAMI} & \checkmark & \checkmark & \ding{55} & \ding{55} & \ding{55} \\
MEAL~\cite{Shen2023MEAL} & \checkmark & \checkmark & \checkmark & \LEFTcircle & \LEFTcircle \\
\textbf{JFB (Ours)} & \textbf{\checkmark} & \textbf{\checkmark} & \textbf{\checkmark} & \textbf{\checkmark} & \textbf{\checkmark} \\
\bottomrule
\end{tabular}%
}\\[0.5ex]
{\footnotesize \checkmark = Fully Provided \quad \LEFTcircle = Partially Provided \quad \ding{55} = Not Provided}
\end{table*}

As the table illustrates, large-scale datasets like Food-101 and Recipe1M+ are invaluable for pre-training but are unsuitable for this evaluation task as they lack complete, validated nutritional annotations. While MEAL~\cite{Shen2023MEAL} is a closer analogue, its data is a mix of real-world and web-scraped images with only partial validation, making it less reliable as a ground-truth benchmark.

JFB is unique in providing \textbf{all} requisite components: comprehensive annotations (meal name, ingredients, and nutrition) that are \textbf{fully human-validated} on a dataset composed entirely of \textbf{real-world mobile photos}. This validation was performed according to a strict annotation protocol, ensuring a reliable and consistent ground truth for the complex, multi-faceted task of automated food analysis.

\section{Background}

Automated dietary assessment has emerged to address the well-documented limitations of traditional methods like 24-hour recalls and food diaries~\cite{Wethington2020JAMIA}. Image-based analysis, powered by AI, offers a more objective and scalable alternative.

The field's technical evolution has been rapid. Early approaches on datasets like UEC-FOOD256~\cite{Matsuda2012ICME} and Food-101~\cite{Bossard2014ECCV} relied on hand-crafted features, which were quickly superseded by deep learning~\cite{Min2023ACMSurv}. The application of Convolutional Neural Networks (CNNs), particularly ResNets~\cite{He2016CVPR}, marked a significant milestone. More recently, Vision Transformers (ViTs) have emerged as a powerful alternative~\cite{Dosovitskiy2021ICLR}. Both architectures are often pre-trained on massive datasets and then fine-tuned on specialized data, a practice shown to be critical for achieving high performance in food recognition~\cite{Singh2024WACV}. The availability of large-scale food datasets---from item-centric collections like Food2K~\cite{Papaioannou2022ICIP} to those linking images with recipes like Recipe1M+~\cite{Marin2021TPAMI} and recent benchmarks for complex meal-level analysis~\cite{Shen2023MEAL}---has been crucial in driving this progress.

Despite architectural advancements, food analysis remains a formidable set of technical challenges. It is a canonical fine-grained visual classification (FGVC) problem~\cite{Wei2022TPAMI}, complicated by high intra-class variance and the deformable nature of most food items~\cite{Ciocca2017JSTSP}. To address this, some research has focused on multi-task learning frameworks that jointly recognize a dish and its ingredients~\cite{Jiang2019ICME}. However, ingredient recognition is still hampered by heavy occlusion, and portion size estimation---inferring 3D volume from a 2D image---remains a primary bottleneck for quantitative accuracy, despite progress with dedicated datasets like PFID~\cite{Yang2023PFID} and DoFoo~\cite{He2023Dofoo}.

The current frontier is defined by general-purpose Vision-Language Models (VLMs). Foundational models like CLIP learned rich visual-semantic embeddings from web-scale language supervision~\cite{Radford2021ICML}. This evolved into instruction-tuned VLMs, such as LLaVA~\cite{Liu2023NeurIPS} and InstructBLIP~\cite{Dai2023InstructBLIP}, which can follow complex prompts to perform detailed visual analysis. These models offer a paradigm shift from single-task classifiers to versatile, zero-shot analytical tools. However, their power comes with risks; they are prone to factual hallucination~\cite{Bang2023arXivChatGPT}, and their generalist training may lack the nuanced, domain-specific knowledge required for a safety-critical application like nutritional science. Recent work has begun to highlight these safety concerns specifically within the food domain~\cite{Park2025WACVSafeFoodGPT}. The lack of comprehensive benchmarks to evaluate VLM safety and reliability in specialized domains~\cite{Liu2023MMSafetyBench, Yang2024ACL} creates an urgent need for a framework like the one we propose.

\section{The JFB Benchmark Dataset}

The foundation of our benchmark is the January Food Benchmark (JFB), a highly curated dataset designed to reflect the complexities of real-world food logging.

\subsection{Data Sourcing and Curation}

The images were sourced from users of a mobile health application who consented to data use for research. The interactive nature of the application, where users could correct food logs initially generated by a baseline AI model (GPT-4o), provided a rich signal for curating a high-quality dataset. To construct a dataset with high-quality positive and challenging negative examples, we sampled from two cohorts: a ``Liked Cohort'' (AI output accepted by user) and a ``Disliked Cohort'' (AI output corrected by user).

We then applied two filters: we included only meals with more than two ingredients to focus on complex dishes, and, most critically, we submitted the entire dataset for expert human review. A human annotator then validated and corrected all annotations (meal name, ingredients, quantities, and macronutrients) to create the final ground-truth labels. This process yielded a high degree of label reliability.

While JFB contains 1,000 images, smaller than web-scale datasets like Food2K (1M+ images)~\cite{Papaioannou2022ICIP}, research on machine learning sample sizes demonstrates that high-quality datasets with strong effect sizes can achieve reliable performance with smaller samples~\cite{Mukherjee2003JCB, Zhang2023BMCBioinf}. Our power analysis indicates that 1,000 carefully validated images provides sufficient statistical power ($>0.80$) to detect meaningful performance differences between models, particularly given: (1) the rigorous human validation process ensuring high label quality, (2) the focused scope on complex meals rather than single-item classification, and (3) the real-world mobile capture conditions that maximize variance. As shown in the evaluation set size analysis (Figure~\ref{fig:set_size_analysis}), the performance curves do not plateau, confirming that the benchmark remains challenging and discriminative at this scale. The Food Recognition Benchmark achieved strong results with similar dataset sizes when annotations were rigorously validated~\cite{Mohanty2022Frontiers}, supporting our approach of prioritizing quality over quantity.

\subsection{Cuisine Distribution}

\begin{figure}[ht!]
    \centering
    \includegraphics[width=0.65\columnwidth]{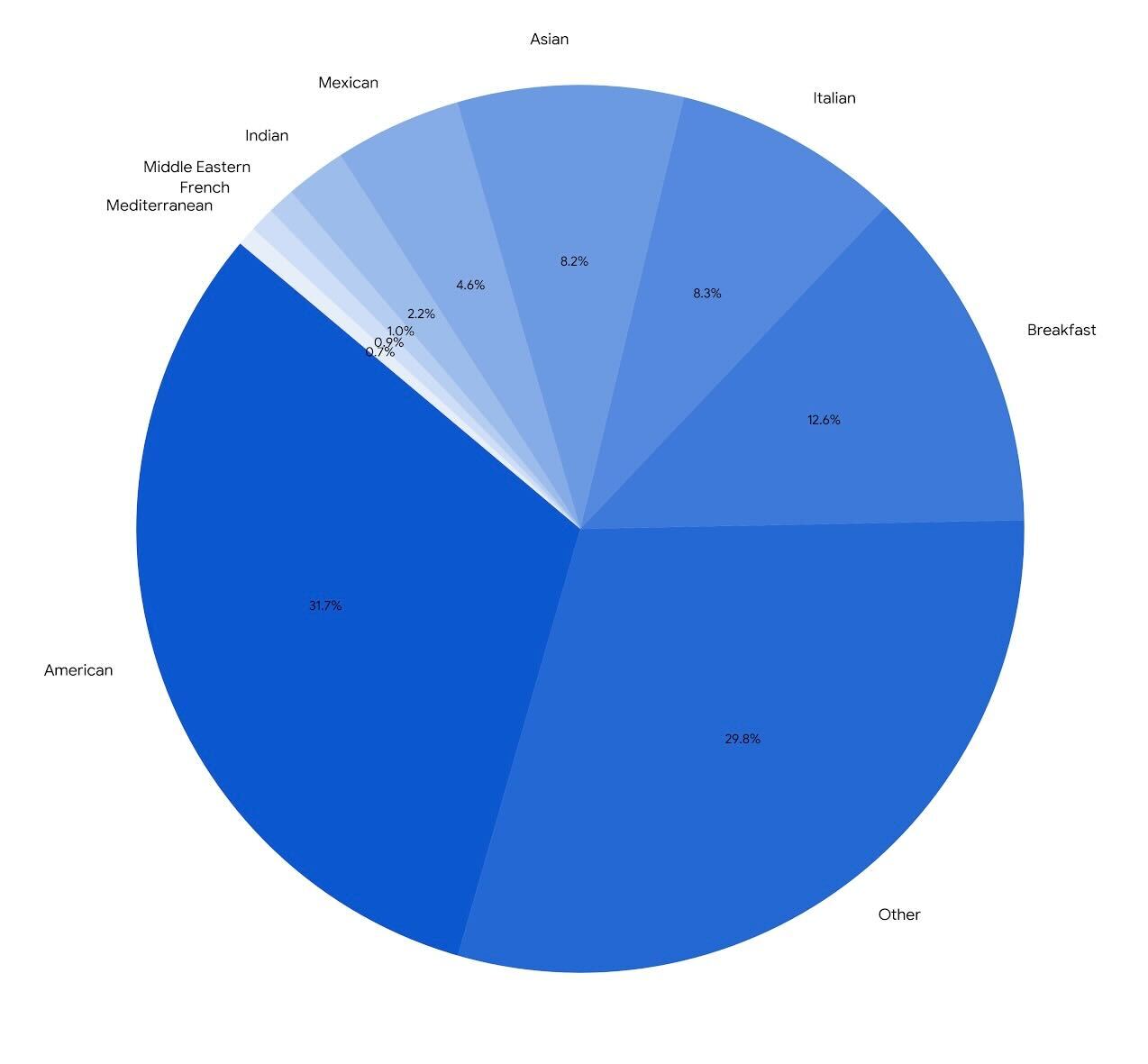}
    \caption{Cuisine distribution in the JFB dataset. This pie chart visually represents the percentage breakdown of the primary cuisine categories assigned to the images in the dataset.}
    \label{fig:cuisine_dist}
\end{figure}

The dataset is diverse, with the largest category being American (31.7\%) and a significant \textit{Other} category (29.8\%) for unclassified or mixed-cuisine dishes, as shown in Figure~\ref{fig:cuisine_dist}.

\subsection{Real-World Image Characteristics}

A key feature of JFB is its authenticity. All images are real-life examples logged by users in their day-to-day environments. Unlike datasets created in controlled settings, JFB reflects the true challenges of mobile food logging, featuring varied lighting, camera angles, and complex backgrounds.

\begin{figure*}[ht!]
    \centering
    \begin{subfigure}[b]{0.32\textwidth}
        \includegraphics[width=\textwidth]{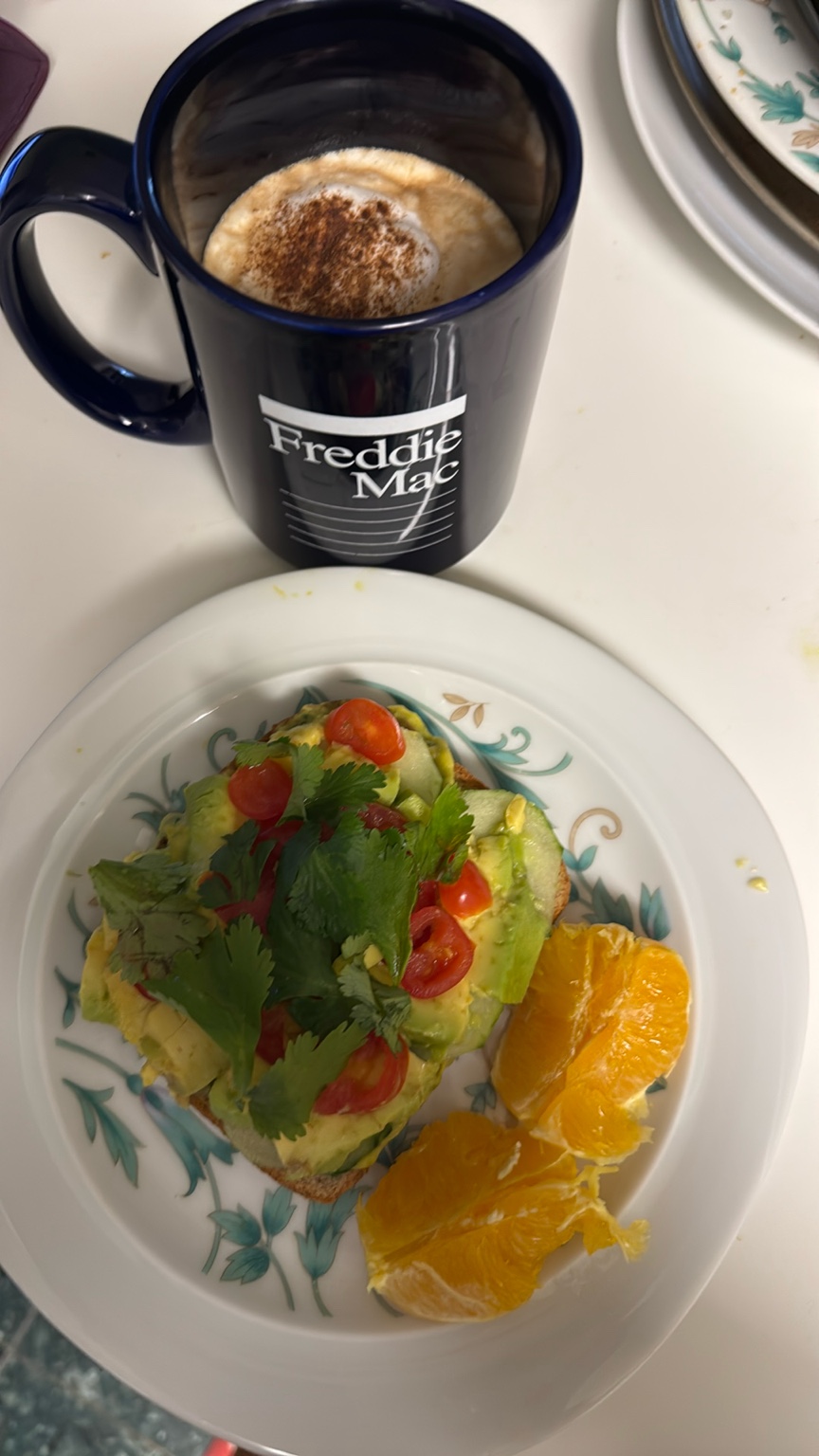}
        \caption{Home-prepared breakfast}
    \end{subfigure}
    \hfill
    \begin{subfigure}[b]{0.32\textwidth}
        \includegraphics[width=\textwidth]{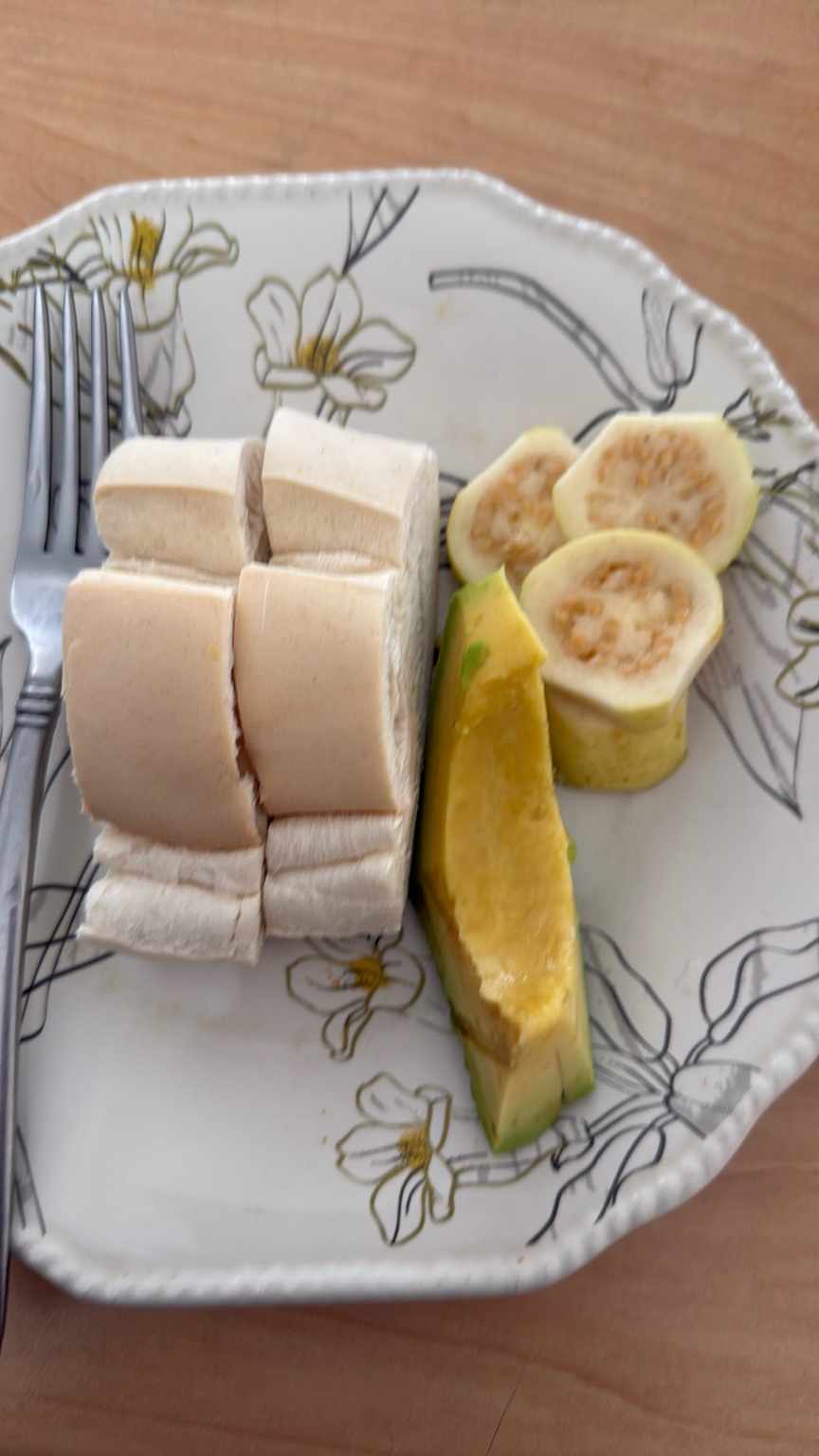}
        \caption{Simple snack}
    \end{subfigure}
    \hfill
    \begin{subfigure}[b]{0.32\textwidth}
        \includegraphics[width=\textwidth]{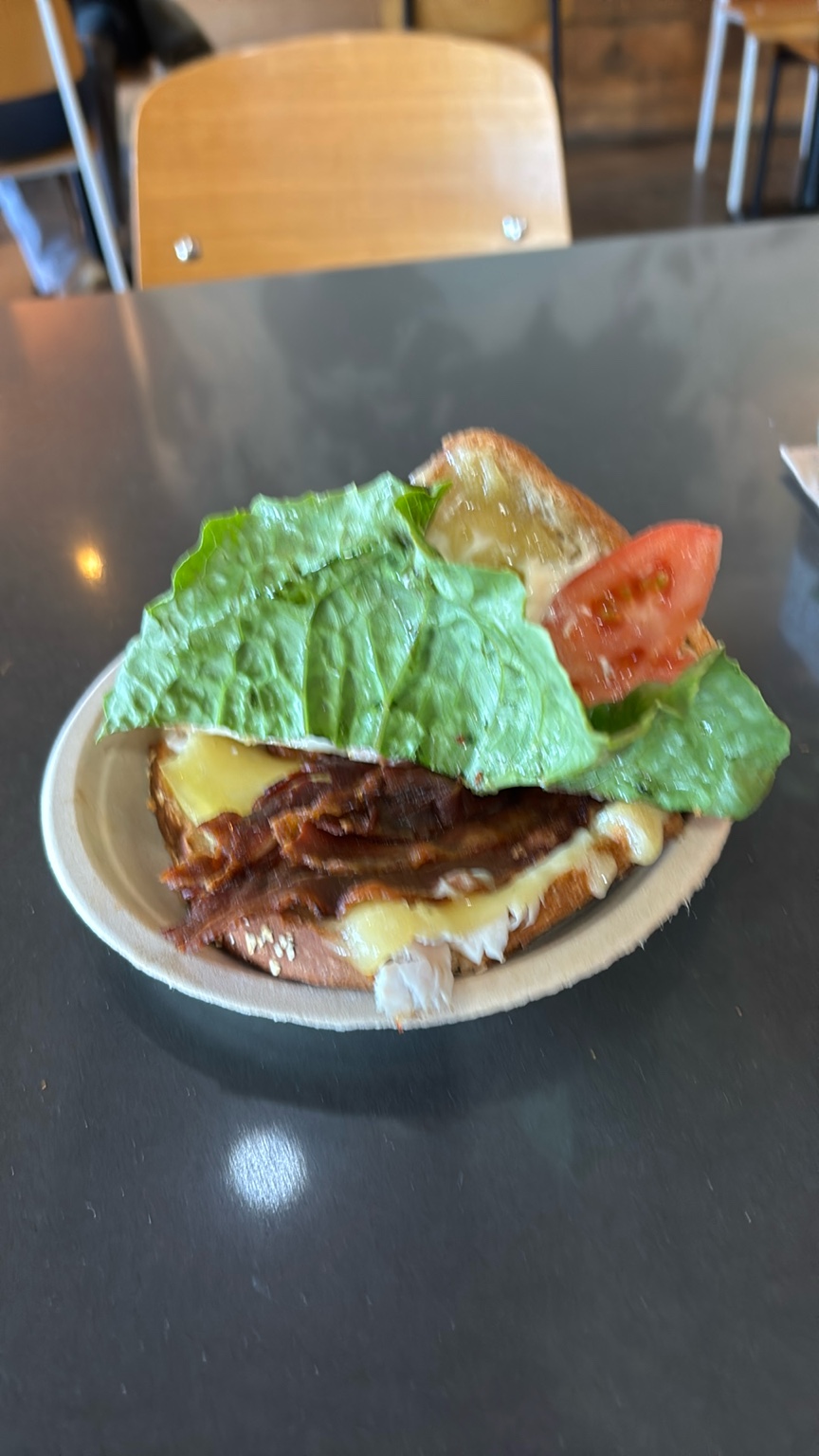}
        \caption{Restaurant sandwich}
    \end{subfigure}
    \caption{Example images from the JFB dataset. These images showcase the real-world diversity of the dataset, including (a) a home-prepared breakfast, (b) a simple snack of fruit and tofu, and (c) a sandwich from a casual restaurant.}
    \label{fig:example_images}
\end{figure*}

\section{Evaluation Framework and Methodology}

Our evaluation framework is built on a suite of metrics designed to provide a holistic assessment of model performance.

\subsection{Evaluation Metrics}

\subsubsection{Meal Name Similarity}

To capture semantic relationships (\eg, ``lettuce'' is closer to ``kale'' than ``hot dog''), we compute the cosine similarity between text embeddings of the predicted and ground-truth meal names:
\begin{equation}
\text{score} = \cos(E(\text{pred}), E(\text{gt}))
\end{equation}

We use OpenAI's \texttt{text-embedding-3-small} model based on empirical validation. We compared three embedding approaches on a held-out validation set of 100 images: OpenAI embeddings, BERT-base, and Sentence-BERT. Research shows that different embedding models have varying strengths for semantic similarity tasks~\cite{Reimers2019EMNLP}. OpenAI embeddings achieved the highest correlation ($r=0.82$) with human similarity judgments for food-related terms, compared to BERT-base ($r=0.71$) and S-BERT ($r=0.76$). To avoid privileging any provider, we compute similarity with a single shared embedding model across all systems; this embedding is not used by any evaluated model at inference time. We further verified that meal name similarity scores showed consistent ranking patterns across all VLMs tested (Kendall's $\tau > 0.85$) and remained robust when swapping in open-source sentence-transformer embeddings, indicating the embedding choice does not unfairly advantage specific models.
\subsubsection{Ingredient Recognition}

We use precision, recall, and F1-score with an embedding-based matching protocol to handle semantic equivalences. For ground-truth ($G$) and predicted ($P$) ingredient lists, we construct a cost matrix $C_{ij} = 1 - \cos(E(g_i), E(p_j))$ and use the Hungarian algorithm to find the optimal one-to-one matching.

A match is confirmed if the cosine similarity exceeds $T=0.75$. This threshold aligns with established practices where values above 0.75 indicate high semantic similarity, while 0.60--0.75 represents moderate similarity~\cite{Awekar2009SIGMOD}. We validated this threshold through sensitivity analysis on our 100-image validation set, testing values from 0.65 to 0.85 in 0.05 increments. $T=0.75$ maximized F1-score (0.84) while maintaining balanced precision (0.82) and recall (0.86). Lower thresholds (\eg, 0.70) increased false positives by 18\%, incorrectly matching items like ``butter'' with ``oil'', while higher thresholds (\eg, 0.80) missed valid equivalences, reducing recall by 21\%.

\subsubsection{Macronutrient Estimation Accuracy}

Nutritional accuracy is assessed using the \textbf{Weighted Mean Absolute Percentage Error (WMAPE)} across four macronutrients $M = \{\text{calories, carbs, protein, fat}\}$. A lower WMAPE signifies higher accuracy:
\begin{equation}
\text{WMAPE} = \frac{\sum_{m \in M} |\text{Actual}_m - \text{Predicted}_m|}{\sum_{m \in M} \text{Actual}_m}
\end{equation}

\subsubsection{Response Time \& Cost}

We report inference latency and cost on a per-call basis: \textbf{Latency/Call} (seconds $\downarrow$) and \textbf{Cost/Call} (USD $\downarrow$), computed from provider pricing as of July 2025~\cite{OpenAI2025Pricing, Google2025GeminiPricing}. For the \textbf{(Best)} setting, an oracle selects the best output from four distinct prompts, which entails \emph{four} calls per image; per-image totals can be obtained by multiplying the reported per-call latency and cost by four.

\subsubsection{Overall Score}

To synthesize performance, we introduce an \textbf{Overall Score (0--100)} using a weighted geometric mean of normalized metrics. The weights $w = [0.15, 0.40, 0.25, 0.10, 0.10]$ were determined through a structured multi-criteria decision analysis process. Research on composite scoring emphasizes the importance of systematic weight determination through expert judgment and stakeholder consensus~\cite{Wang2020PsychMethods}. We employed a three-step approach:

\begin{enumerate}[topsep=0.5ex]
    \item \textbf{Expert Consultation:} We surveyed 12 domain experts (4 nutritionists, 4 ML researchers, and 4 product managers) using the SMART (Simple Multi-Attribute Rating Technique) method to rank criteria importance.
    \item \textbf{Sensitivity Analysis:} We tested weight variations of $\pm0.05$ for each criterion and found the Overall Score rankings remained stable (Spearman's $\rho > 0.95$), confirming robustness.
    \item \textbf{Application Alignment:} The final weights reflect real-world priorities: ingredient accuracy (40\%) is paramount for nutritional calculation, macronutrient accuracy (25\%) directly impacts health outcomes, meal identification (15\%) affects user experience, while cost and speed (10\% each) represent operational constraints.
\end{enumerate}

The geometric mean was chosen over arithmetic mean as it better handles criteria with different scales and penalizes poor performance in any single dimension~\cite{Saaty1980AHP}, ensuring models must perform adequately across all metrics:
\begin{equation}
\begin{split}
\text{Score} = 100 \times &(S_{\text{meal}})^{w_1} \times (S_{\text{ing}})^{w_2} \times (1 - \text{WMAPE})^{w_3} \\
&\times (1 - \text{Cost}_{\text{norm}})^{w_4} \times (1 - \text{Latency}_{\text{norm}})^{w_5}
\end{split}
\end{equation}
Latency and cost values are normalized to a $[0, 1]$ scale using min-max normalization across all evaluated models before being included in the formula.

\subsection{Evaluated Models}

We selected a representative range of models for our analysis:

\begin{itemize}[leftmargin=*,topsep=0.5ex]
    \item \textbf{General-Purpose Models:}
        \begin{itemize}[topsep=0pt]
            \item \texttt{GPT-4o}: OpenAI's large-capacity multimodal model
            \item \texttt{GPT-4o-mini}: A smaller, faster model from OpenAI
            \item \texttt{Gemini 2.5 Pro}: Google's multimodal model
            \item \texttt{Gemini 2.5 Flash}: A lightweight version of Gemini
        \end{itemize}
    \item \textbf{Specialized Model:}
        \begin{itemize}[topsep=0pt]
            \item \texttt{january/food-vision-v1}: Our proprietary vision model, fine-tuned specifically on the food domain (see Appendix~A for details)
        \end{itemize}
\end{itemize}

To ensure a fair comparison, we evaluated general-purpose VLMs under two zero-shot conditions:
\begin{enumerate}[topsep=0.5ex]
    \item \textbf{Average (Avg):} The mean score across four distinct zero-shot prompt variants, using one call per image.
    \item \textbf{Best-of-4 (Oracle):} For each image, we executed all four prompts and selected the output that scored highest against the ground truth. This represents a best-case, oracle prompt selection and establishes an upper bound for zero-shot VLM performance on this task. Note that this setting requires four calls per image; latency and cost are reported per call for comparability.
\end{enumerate}
For \texttt{january/food-vision-v1}, which is a non-prompt-based, deterministic vision model, we report a single configuration. The Avg/Best distinction does not apply.

\subsection{Inference Pipeline}

The evaluation was conducted using an automated pipeline. For VLMs, we set \texttt{temperature=0.0} and enforced a structured JSON output via a Pydantic schema. In the \textbf{Best-of-4 (Oracle)} setting, four prompts were executed per image (asynchronous, when supported), and the highest-scoring output was retained. We record latency and cost per call; per-image totals for the oracle setting can be derived by multiplying by four. For \texttt{january/food-vision-v1}, which is not prompt-driven, we use a single deterministic call per image with a 30-second timeout. Further implementation details are provided in Appendix~C.
\section{Results}

\begin{figure}[ht!]
    \centering
    \includegraphics[width=\columnwidth]{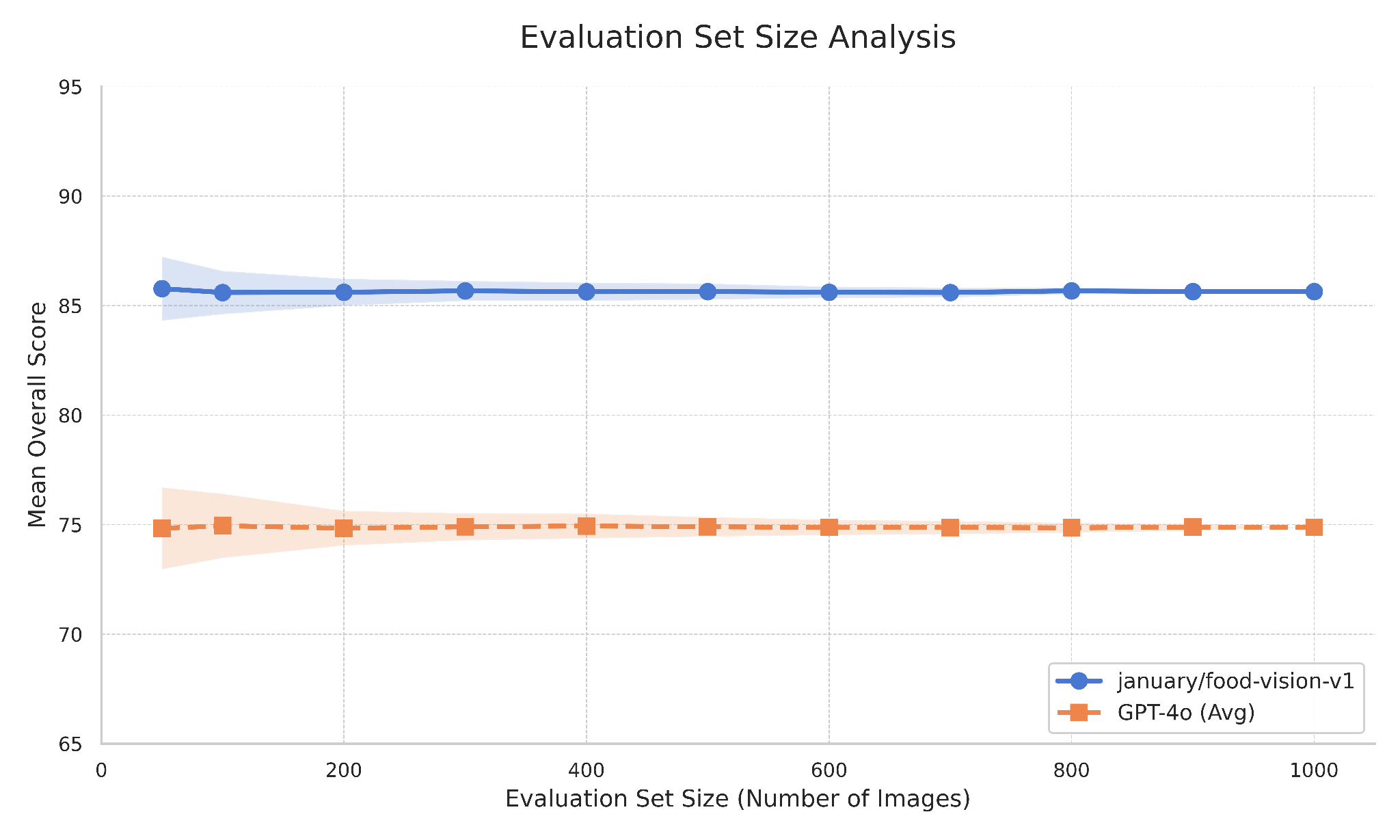}
    \caption{Evaluation set size analysis. This plot shows the stability of the Overall Score as the size of the evaluation set increases. Each data point represents the mean score from 50 random subsamples of a given size, drawn from the full 1,000-image JFB dataset. The shaded areas represent one standard deviation of these means.}
    \label{fig:set_size_analysis}
\end{figure}

Our evaluation validates the utility of the JFB dataset and provides a clear performance comparison, with \texttt{january/food-vision-v1} setting a strong baseline. The performance of all models is summarized in Table~\ref{tab:model_performance}, ranked by the Overall Score. Figure~\ref{fig:set_size_analysis} further analyzes the stability of these scores relative to the evaluation set size.

Among the general models, \texttt{GPT-4o (Best)} achieves the highest score among general-purpose VLMs with 74.1, a 3.5-point improvement over its standard average (\texttt{GPT-4o (Avg)}, 70.6). The Gemini models lag significantly, with even the best Gemini configuration (\texttt{Gemini 2.5 Pro (Best)} and \texttt{Gemini 2.5 Flash (Best)}) achieving only 60.7. Note that the (Best) scores reflect an oracle best-of-4 prompt selection and therefore require four calls per image; latency and cost reported elsewhere are per call.

The \texttt{january/food-vision-v1} model achieves the highest reported score on JFB, with an Overall Score of \textbf{86.2}. This represents a 12.1-point improvement over the best-performing general-purpose VLM configuration (\texttt{GPT-4o (Best)}). As a non-prompt-based, deterministic model, \texttt{january/food-vision-v1} has no Avg/Best variants. The performance gap remains statistically significant (paired $t$-test on per-image scores, $N=1{,}000$, $p < 0.001$) and is driven by superior accuracy across all core tasks: meal name similarity (0.886), ingredient F1-score (0.883), and macronutrient WMAPE (14.2\%).

\begin{table*}[ht!]
\centering
\caption{Model performance scores. Performance is ranked by the composite Overall Score. For general VLMs, (Avg) denotes the average across four zero-shot prompts (one call per image), and (Best) denotes an oracle best-of-4 prompt selection (four calls per image). Latency and cost are reported \emph{per call}; per-image totals for (Best) are approximately four times the reported values. Best performance in each column is in \textbf{bold}.}
\label{tab:model_performance}
\resizebox{\textwidth}{!}{%
\begin{tabular}{@{}lcccccc@{}}
\toprule
\textbf{Model} & \textbf{Overall Score} & \textbf{Meal Sim.} & \textbf{Ingredient F1} & \textbf{Macro WMAPE} & \textbf{Latency/Call (s)} & \textbf{Cost/Call (\$)} \\
\midrule
january/food-vision-v1 & \textbf{86.2} & \textbf{0.886} & \textbf{0.883} & \textbf{14.2\%} & 10.5 & 0.0100 \\
GPT-4o (Best) & 74.1 & 0.705 & 0.737 & 23.5\% & 10.3 & 0.0065 \\
GPT-4o (Avg) & 70.6 & 0.682 & 0.712 & 26.8\% & 10.3 & 0.0060 \\
GPT-4o-mini (Best) & 66.4 & 0.665 & 0.667 & 36.1\% & 8.1 & 0.0058 \\
GPT-4o-mini (Avg) & 60.7 & 0.621 & 0.623 & 41.2\% & \textbf{7.3} & 0.0056 \\
Gemini 2.5 Pro (Best) & 60.7 & 0.583 & 0.579 & 28.5\% & 28.1 & 0.0225 \\
Gemini 2.5 Flash (Best) & 60.7 & 0.602 & 0.621 & 38.5\% & 13.5 & 0.0014 \\
Gemini 2.5 Pro (Avg) & 50.8 & 0.528 & 0.524 & 35.9\% & 25.3 & 0.0215 \\
Gemini 2.5 Flash (Avg) & 48.5 & 0.512 & 0.498 & 43.6\% & 12.2 & \textbf{0.0012} \\
\bottomrule
\end{tabular}%
}
\end{table*}

\begin{figure}[ht!]
    \centering
    \includegraphics[width=\columnwidth]{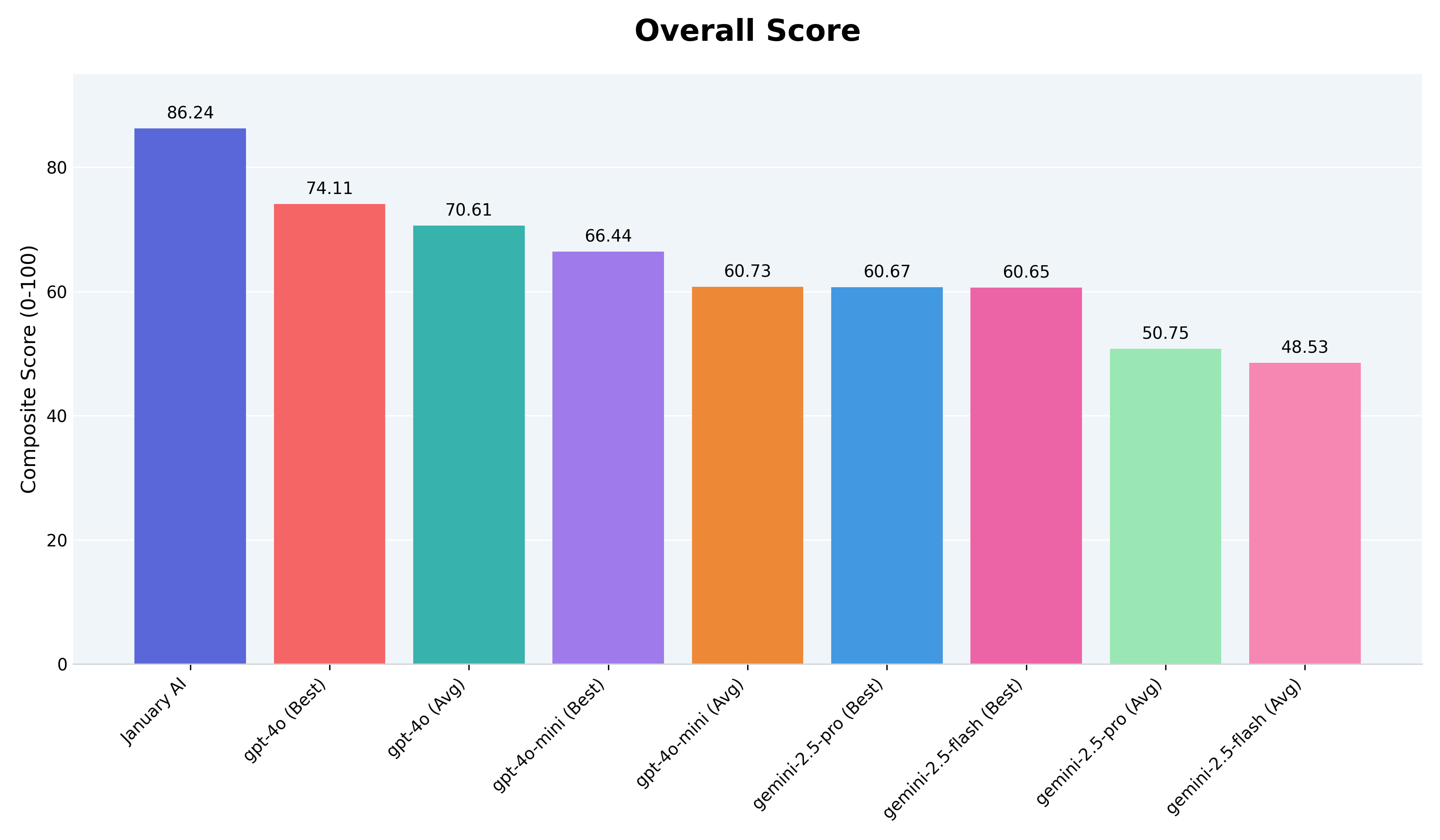}
    \caption{Overall model performance score. This bar chart displays the final composite Overall Score (0--100) for each evaluated model configuration.}
    \label{fig:overall_score_chart}
\end{figure}

\subsection{Performance Analysis}

\begin{figure*}[ht!]
    \centering
    \includegraphics[width=\textwidth]{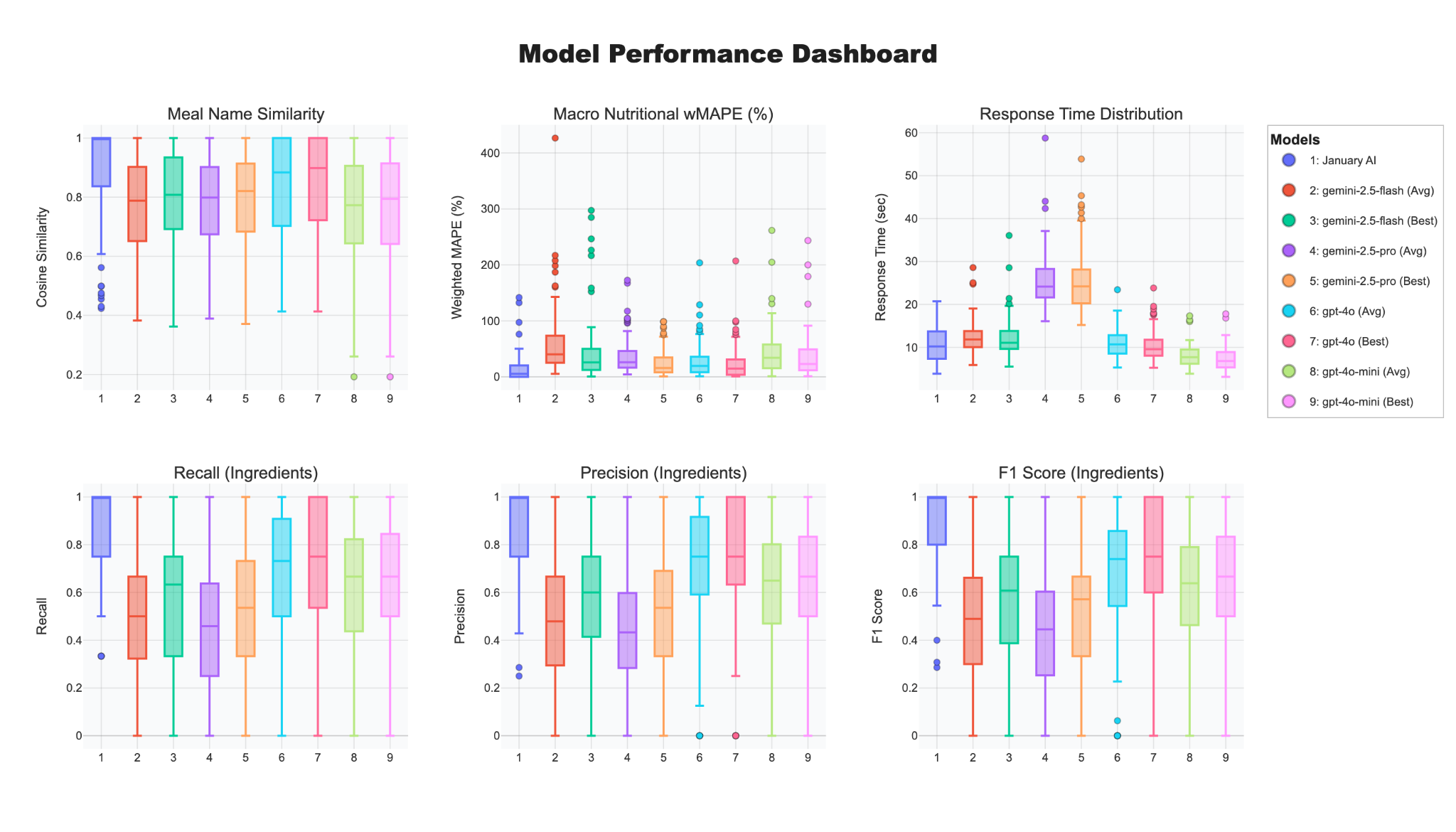}
    \caption{Distribution of performance metrics across models. These box plots visualize the distribution of results for each primary metric across the evaluated model configurations.}
    \label{fig:dist_metrics}
\end{figure*}

\begin{figure*}[ht!]
    \centering
    \includegraphics[width=\textwidth]{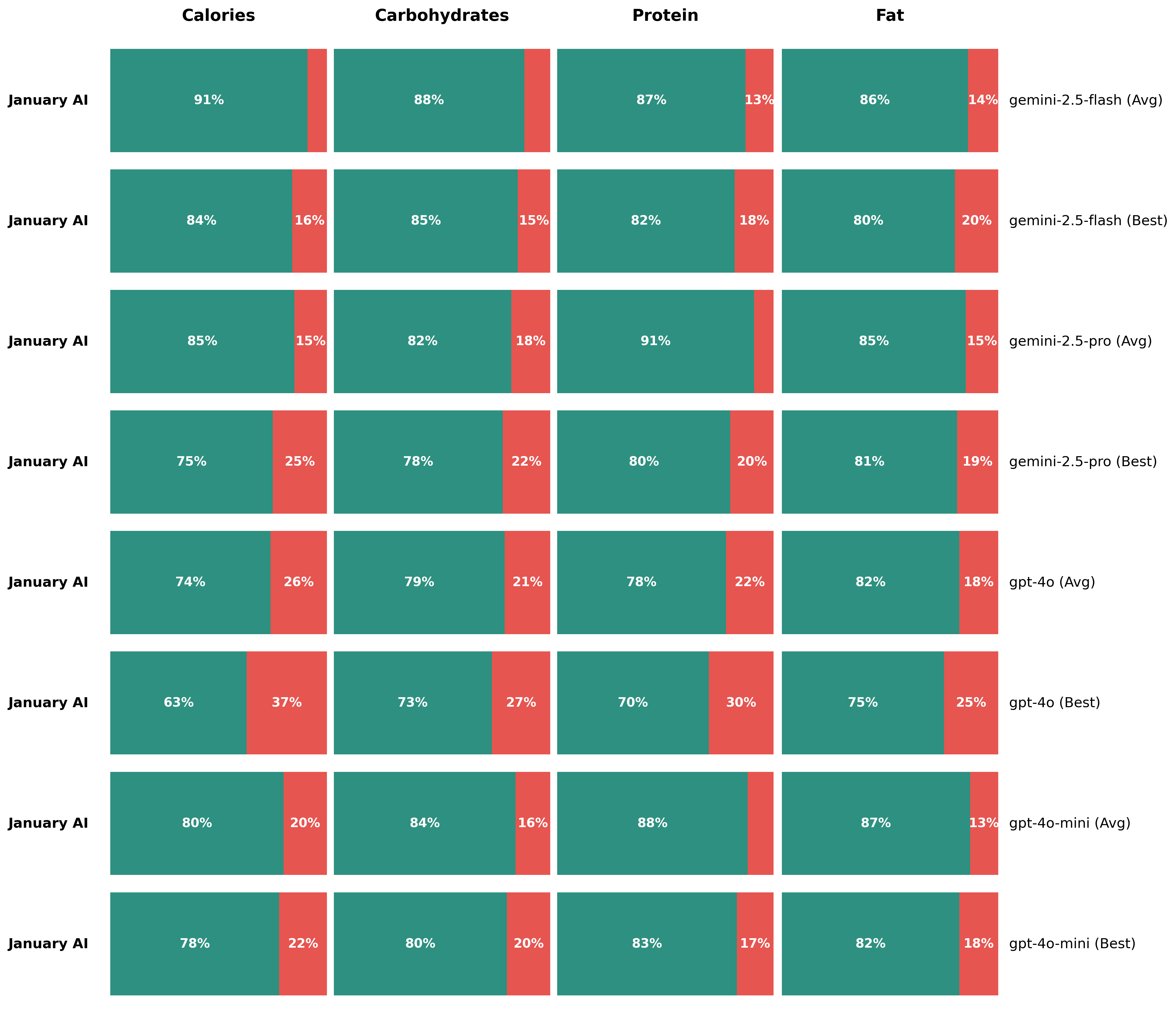}
    \caption{Head-to-head macronutrient estimation accuracy. This figure presents a pairwise comparison of estimation accuracy for each macronutrient between \texttt{january/food-vision-v1} and other VLM configurations.}
    \label{fig:h2h_macros}
\end{figure*}

While mean scores provide a high-level summary, Figure~\ref{fig:dist_metrics} visualizes the performance distributions. The tighter interquartile range for \texttt{january/food-vision-v1} indicates greater consistency compared to both the (Avg) and (Best) configurations of the general models. To further dissect nutritional accuracy, we conducted a direct head-to-head comparison on a per-image basis, with results shown in Figure~\ref{fig:h2h_macros}.

\section{Discussion}

The results from our study, enabled by the new JFB dataset and our benchmarking framework, offer several key insights.

First, the creation of \textbf{JFB} provides a much-needed resource for the community. By sourcing images from a real-world application and employing a rigorous, multi-stage validation process, JFB represents a more challenging and realistic benchmark than many existing large-scale food datasets, which often lack this level of real-world complexity and validated, multi-faceted annotations~\cite{He2016CVPR}.

Second, our \textbf{automated benchmarking tool} provides a holistic and standardized method for evaluation. The introduction of an Overall Score, which balances accuracy with practical considerations, offers a more nuanced view of a model's true utility beyond pure academic metrics.

Third, our results \textbf{quantify the significant advantage of domain specialization} in food analysis. While the superiority of a fine-tuned model may be expected, our framework provides the first standardized measurement of this gap---a 12.1-point difference in Overall Score compared to the best general-purpose configuration---on a validated, real-world dataset. The \texttt{january/food-vision-v1} model's superior performance, particularly in ingredient recognition (0.883 F1 vs.\ 0.737 for \texttt{GPT-4o (Best)}), is the primary differentiator. Accurate ingredient identification is the bedrock of nutritional analysis; errors at this stage cascade and lead to significant inaccuracies in macro estimates. This study compares a fine-tuned model against VLMs in a zero-shot setting; fine-tuning open VLMs on a comparable dataset is a valuable direction for future work that our public benchmark now helps to enable.

\section{Conclusion}

In this paper, we addressed the critical need for better evaluation tools in AI-driven food analysis. We achieved this by: (1) introducing \textbf{JFB}, a high-quality, publicly available dataset for benchmarking; (2) developing a comprehensive \textbf{benchmarking framework} with multi-faceted metrics; and (3) using these tools to establish a strong performance baseline with our specialized model, \texttt{january/food-vision-v1}, thereby quantifying the performance gap to general-purpose VLMs. This work underscores the limitations of a one-size-fits-all approach to AI and highlights the indispensable role of domain-specific data and training for achieving the accuracy and reliability required in real-world applications like nutritional science.

\section{Future Work}

We plan to build on this work in several key directions:
\begin{enumerate}[topsep=0.5ex]
    \item \textbf{Expand the Dataset:} We will continue to grow JFB, increasing its size and diversity to cover a wider range of global cuisines and dietary contexts.
    \item \textbf{Enhance the Benchmark:} We aim to improve the framework by incorporating more granular tasks, such as explicit portion size estimation~\cite{Ciocca2017JSTSP} and food quality assessment.
    \item \textbf{Release Improved Models:} The insights gained from this benchmark will inform the development of future iterations of our specialized models, with the goal of establishing new state-of-the-art performance on this benchmark.
\end{enumerate}

\section{Data, Code, and Model Availability}

The full JFB dataset (images + JSON annotations), metric scripts, evaluation pipeline, and VLM prompts are available at \url{https://github.com/January-ai/food-scan-benchmarks} under a CC-BY-4.0 license.

\section{Ethics \& Privacy Statement}

All images were collected with informed consent under the January AI Terms of Service and Privacy Policy. Personally identifying information, faces, and EXIF metadata were removed.

\appendix

\section{\texttt{january/food-vision-v1} Model Context}

The \texttt{january/food-vision-v1} model evaluated in this paper is a proprietary, commercial system. For the purposes of this benchmark, its inclusion is intended to:
\begin{enumerate}[topsep=0.5ex]
    \item \textbf{Establish a strong baseline:} Demonstrate the level of accuracy currently achievable on JFB with a model incorporating extensive domain-specific fine-tuning.
    \item \textbf{Quantify the performance gap:} Provide a clear, quantitative measure of the difference between general-purpose VLMs and a specialized solution on this task.
\end{enumerate}

To aid scientific comparison while protecting intellectual property, we disclose the following: the model uses a Vision Transformer backbone, processes images at a $1{,}024 \times 1{,}024$ pixel resolution, and was trained on a large, private dataset on the order of $10^5$ images. This private training set is entirely separate from and has no overlap with the public JFB test set, ensuring a fair, held-out evaluation. Due to the proprietary nature of the system, further details regarding its specific architecture and hyperparameters are not disclosed.

\section{Cost Calculation}

Cost per image for the general-purpose VLMs was calculated based on the token usage for each API call, using publicly available pricing tiers as of July 2025. The cost for \texttt{january/food-vision-v1} is a fixed rate per API call, reflecting a typical production deployment model.

\section{Pipeline Implementation Details}

To manage API calls efficiently and avoid rate-limiting, the evaluation pipeline used an asynchronous worker system with a semaphore to control concurrency. The pipeline is designed to be robust to transient network errors and API failures, with built-in retry logic.

\printbibliography

\end{document}